\documentclass[letterpaper, 10 pt, conference]{ieeeconf}  

\IEEEoverridecommandlockouts                              

\usepackage[sorting=none]{biblatex} 
\usepackage{graphicx}
\usepackage{gensymb}
\usepackage{nicematrix}
\usepackage{mathtools}
\usepackage[ruled,vlined]{algorithm2e}
\usepackage{verbatim}
\usepackage{multirow}
\usepackage{nccmath}

\title{\LARGE \bf
Canonical mapping as a general-purpose object descriptor for robotic manipulation
}

\author{Benjamin Joffe and Konrad Ahlin
\thanks{All authors are with the Georgia Tech Research Institute,
        Georgia Institute of Technology, Atlanta, GA 30332.
        {\tt\small joffe@gatech.edu}, {\tt\small konrad.ahlin@gtri.gatech.edu}}%
}

\begin{document}

\maketitle
\thispagestyle{empty}
\pagestyle{empty}

\begin{abstract}

Perception is an essential part of robotic manipulation in a semi-structured environment. Traditional approaches produce a narrow task-specific prediction (e.g., object's 6D pose), that cannot be adapted to other tasks and is ill-suited for deformable objects. In this paper, we propose using canonical mapping as a near-universal and flexible object descriptor. We demonstrate that common object representations can be derived from a single pre-trained canonical mapping model, which in turn can be generated with minimal manual effort using an automated data generation and training pipeline. We perform a multi-stage experiment using two robot arms that demonstrate the robustness of the perception approach and the ways it can inform the manipulation strategy, thus serving as a powerful foundation for general-purpose robotic manipulation.

\end{abstract}

\section{INTRODUCTION}

It can be argued that the next frontier for wider robotic adoption lies in semi-structured environments. In such settings, the robot knows what categories of objects it can expect and their general location (e.g., within the workspace); however, their exact appearance, pose, and configuration are not known a priori. Perception algorithms play a key role in generating a suitable and robust representation of such objects to enable a robot to manipulate them. 6D pose is one of the most common and powerful ways to describe an object to a robot. There have been steady improvements in RGB- and RGB-D-based 6D pose estimation \cite{xiang2018posecnn, wang2019densefusion, He_2020_CVPR}, but, generally, they are not sufficiently robust to generalize to arbitrary object instances. The generalization problem is being addressed by the subfield of category-level pose estimation \cite{Wang_2019_CVPR, gsnet2020ke, chen2020category}. However, the 6D pose has a fundamental assumption of the rigidity of the object, while many of the future robotic applications, such as agriculture, food processing, textile, etc, involve the manipulation of \mbox{(semi-)} deformable and articulating objects. 

If the object has articulated parts (e.g., most animals and many manufactured items), the pose estimation model needs to either ignore those parts and still robustly predict the pose of the main body or, additionally, learn a pose estimation model for each part of the object. If the object is semi-deformable, such as soft toys or food items that may change in volume or have soft surfaces but generally preserve the overall shape, 6D Pose can only be usable with a high error tolerance. As demonstrated in our experiments, successful manipulation of such objects can also be achieved by focusing on their surface areas or parts, but, again, traditional approaches would require learning a model for each such part. Finally, a fully deformable object, such as a textile, does not have a standard descriptor; however, it is still suitable for our proposed approach, provided some identifiable visual features exist on its surface.

\begin{figure}[t]
    \centering
    \includegraphics[width=\linewidth]{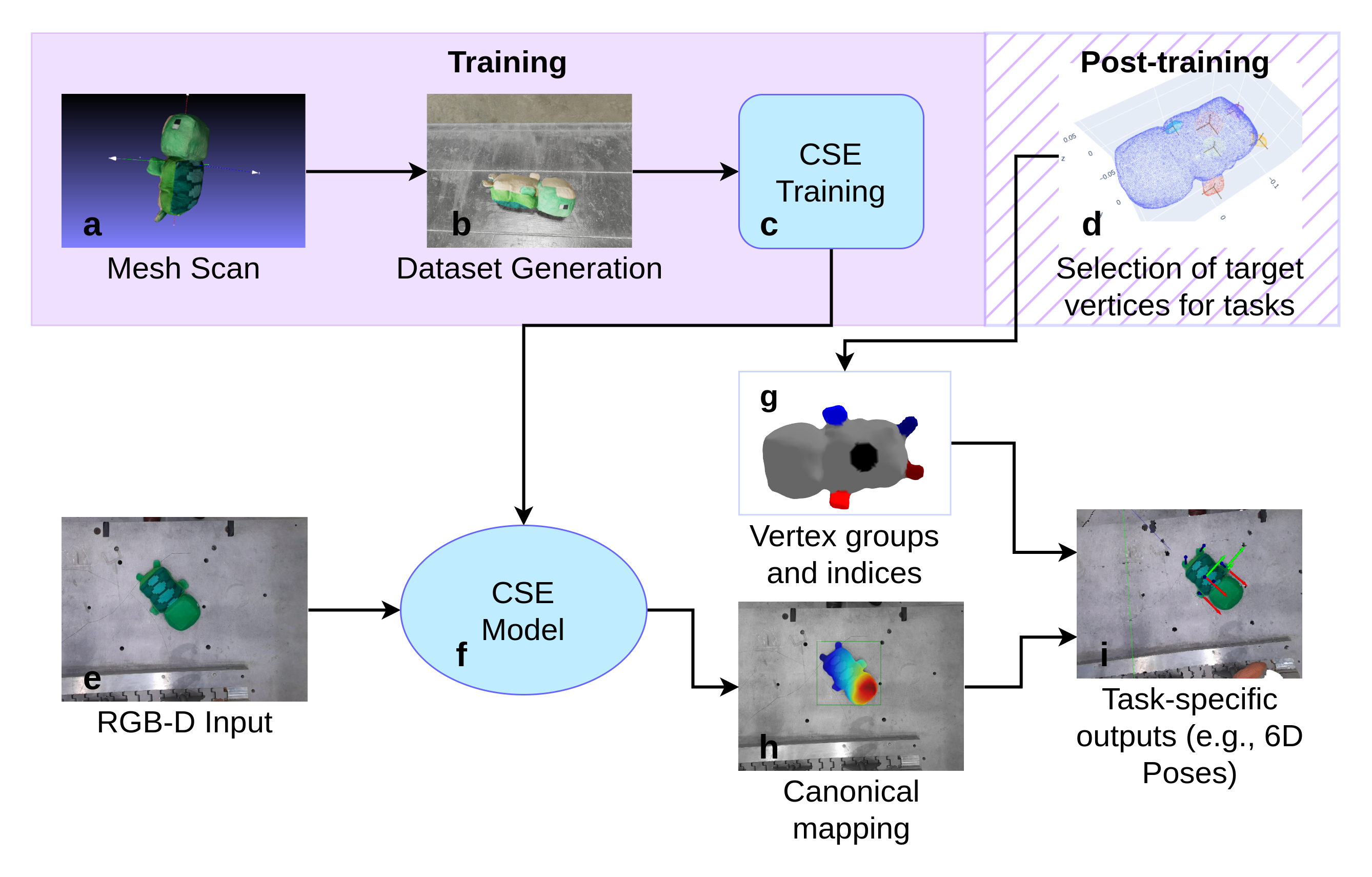}
    \caption{Overview of the approach. Train time (top): (a) Scan of the physical object using photogrammetry app and mesh generation, (b) Synthetic dataset generation, (c) Canonical Mapping model training, (d) Optional task-specific selection of vertices (or vertex groups) for localization or to predict parts' 6D poses. Test time (bottom): (e) given an RGB-D image, (h) predict canonical mapping and use the dense correspondences to estimate (i) the object's rigid 6D pose and an arbitrary number of part 6D poses.}
    \label{fig:cm_overview}
\end{figure}

To address the aforementioned issues of describing a non-rigid object and lack of task adaptability, we propose canonical mapping as a single general-purpose descriptor that can in turn be adapted to an arbitrary task-specific representation required by the robot. Canonical mapping semantically identifies each object's pixel on the image by mapping it to a dense 3D mesh of the object's category. Even if the object is deformed or articulated, the points from those areas are mapped to their fixed locations on the unarticulated canonical 3D mesh. By observing dense correspondences between canonical 3D points and observed points, it's possible to infer the pose of the object as a whole as well as any of its parts (defined broadly as any group of vertices). Since points or areas of interest are defined with respect to the mesh, they don't need to be defined prior or during the training. Therefore, the canonical mapping model needs to be trained only once, and subsequently, the conversion to a task-specific representation can be added at runtime.

Training canonical mapping models requires a substantial amount of annotations, particularly a high-fidelity 3D mesh and matches between keypoints on the image and their corresponding vertex indices. We believe that the proposed object descriptor's universality claim should also extend into the ease of its generation and not just the possibilities of its application. Thus, we developed an automated pipeline to generate the descriptor for a physical object in hand (see  Fig. \ref{fig:cm_overview}). This automation is possible due to (1) availability of off-the-shelf smartphone photogrammetry apps that scan the object and generate suitable high-fidelity 3D mesh in a matter of minutes and (2) photorealistic rendering and simulation environments for synthetic dataset generation. Once the pipeline is set up, the only manual step done by the user that is required to learn the canonical mapping model is performing the scan of a new object and transferring of the mesh. The same mesh is used to render the dataset, generate the required annotations, and train the model. Separate from training, we provide an interactive interface where the user may select individual vertices or vertex groups in order to localize or compute 6D poses of object parts.

We validate our approach by applying the pipeline to a soft toy in order to generate its mesh and learn the canonical mapping model associated with it. We then perform experiments in robot manipulation of the toy to demonstrate the accuracy and applicability of the learned descriptor. In summary, the main contributions of this paper are: (1) demonstrate that a canonical mapping model is a general-purpose descriptor of an object (even deformable and/or articulating), (2) present an automated data generation and training pipeline to produce such a model where human labor is on the order of minutes, (3) show that it enables flexible robotic manipulation.

\section{RELATED WORK}

\subsection{6D Pose Estimation}
6D Pose is perhaps the most common object descriptor used in Robotics. It is also the representation to which we most often convert our predicted descriptor (canonical mapping), as it can be naturally integrated into the manipulation approach. 
6D Pose Estimation is a well-studied problem in Robotics and Computer Vision with methods available for RGB \cite{suwajanakorn2018discovery, tekin2018real, sundermeyer2018implicit} and RGB combined with depth or pointcloud inputs \cite{krull2015learning, kehl2016deep, wang2019densefusion}. Most RGB-based methods rely on the detection of semantic keypoints in the image and solving Perspective-n-Point (PnP) problem to estimate the camera pose \cite{tekin2018real}. Other methods use regression to predict 3D outputs \cite{xiang2018posecnn} or compute embeddings in order to estimate 2D-3D correspondences \cite{haugaard2022surfemb}. Since most traditinoal approaches require labor-intensive semantic keypoint annotations, the usage of synthetic datasets is being more widely adopted. SurfEmb \cite{haugaard2022surfemb} achieves state-of-the-art performance in pose estimation while training purely on synthetic data. 

Until recently, most methods and standard benchmark datasets (e.g., \cite{calli2015benchmarking}) focused on instance-level pose estimation. In such a setting, the model is trained for a specific example of an object and will not generalize to other examples in the object's category. To address this, recent approaches and benchmarks were developed for category-level pose estimation \cite{Wang_2019_CVPR, gsnet2020ke, chen2020category, lin2022single}, where a model trained for a given category (e.g., car, chair, mug) can be applied to novel object instances of that class. While this improves the generalization potential, it is still limited to rigid objects and, as previously discussed, the 6D Pose Estimation methods have very limited application for articulated and deformable objects.

\subsection{Canonical Mapping}

Canonical Mapping refers to the prediction of correspondences between pixels belonging to an observed object and the vertices of its reference (canonical) 3D mesh. The problem can be viewed as an example of a more general area of estimation of dense semantic correspondences. The latter area includes more established tasks, such as computing pixel-to-pixel correspondences (e.g., SIFT). The advantage of using a dense mesh as one of the mapping domains is that it contains a complete representation of the object (rather than a partial view), includes 3D information, and provides a convenient modality to select and define manipulation areas and strategy. To estimate canonical mapping, DensePose \cite{guler2018densepose} partitions a 3D shape into surface parts (parameterized as 2D charts) and learns to regress to those parts' $uv$ coordinates (pixel-to-surface correspondences). Training such a model requires millions of annotated correspondences between images and 3D models. Canonical Surface Mapping \cite{kulkarni2019canonical} approach demonstrated the ability to learn dense pixel-to-3D correspondences with limited supervision by ensuring that the projection of a predicted 3D point back onto the image maps close to the original pixel (cycle of geometric consistency). The work was extended in \cite{kulkarni2020articulation} to the prediction of articulations of predefined object parts (parametrized as 6D poses of vertex groups). Continuous Surface Embeddings (CSE) \cite{neverova2020continuous} method used as part of our approach, addresses computing dense correspondences by learning to predict, for each foreground pixel in an image, an embedding vector of the corresponding vertex in the object's mesh. CSE achieves state-of-the-art performance while using fewer annotations and conceptually simpler architecture. Additionally, the method shows the generalizability potential and the ability to relate and share computation for several object categories.


\section{APPROACH}

\begin{figure*}[t]
    \centering
    \vspace{6pt}
    \includegraphics[width=0.8\textwidth]{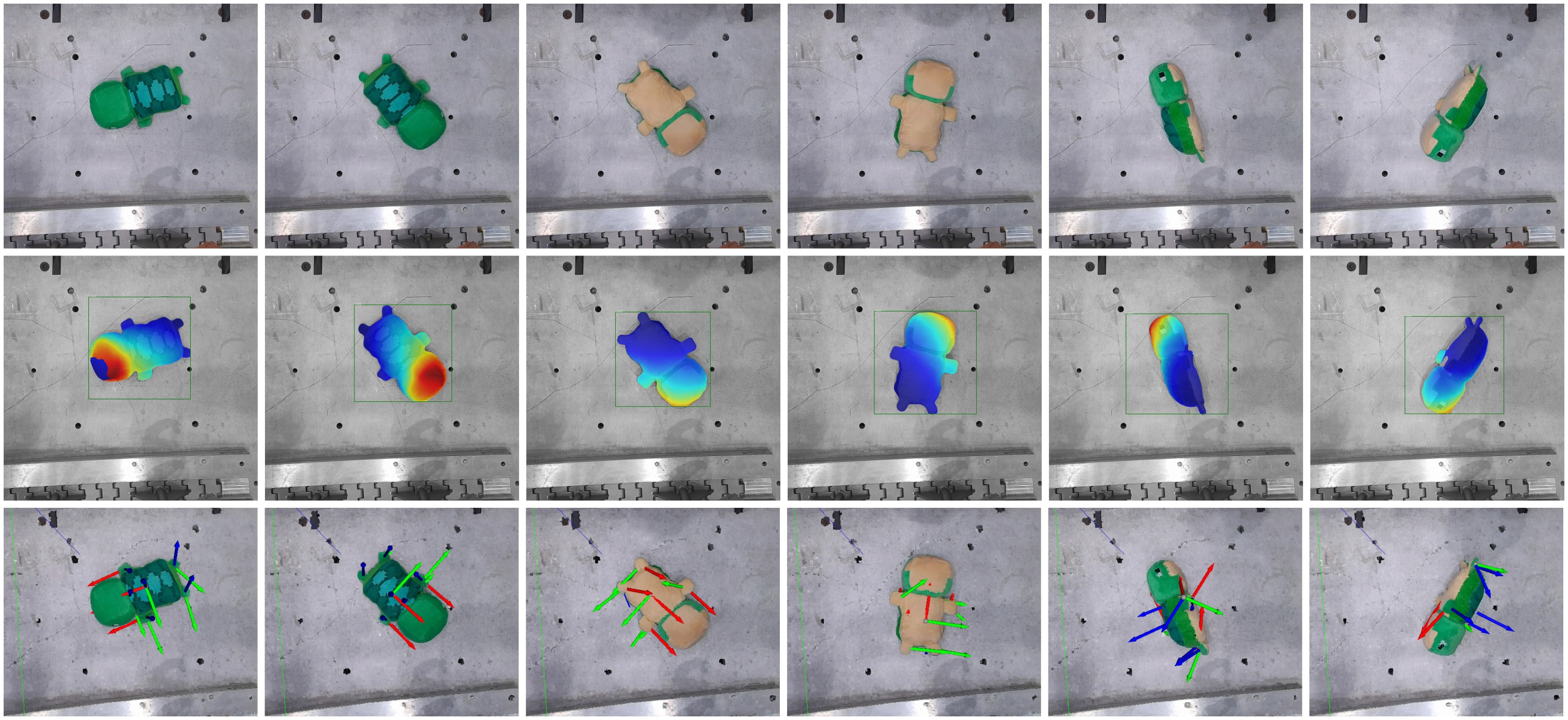}
    \caption{Qualitative results on real images (cropped for legibility). Row 1: input images from Kinect Azure camera. Row 2: predicted canonical mapping visualized using a fixed color map (notice that the color of physical points on the object stays consistent for all orientations). Row 3: 6D Poses for full (rigid) object, and 6 parts (back, belly, left hand, right hand, left foot, right foot), extracted using Algorithm \ref{algo:task_conversion}. Note, that no real images or explicit pose annotations were used for the training or development of the approach.}
    \label{fig:test_results}
\end{figure*}

\subsection{Overview}
Our approach focuses on producing a flexible and robust model for a novel object using a minimal amount of labor (see Fig. \ref{fig:cm_overview}). As such, we avoid extensive data collection and manual annotation and instead rely on automated mesh scanning and synthetic dataset generation. The mesh scanning is done using an off-the-shelf photogrammetry smartphone application. The application performs Structure-from-Motion reconstruction of the object based on about 200 images from varying angles and directly outputs a colored 3D mesh. To provide full coverage and avoid any mesh clean-up, the object was hung on a fishing line. This mesh serves 2 purposes in our pipeline: (1) a canonical mesh — a reference representation of the object, with respect to which desired manipulations are defined, and (2) a high-fidelity mesh used to render the object in realistic scenes using a simulation environment. In conjunction with rendering, the canonical mesh is used to automatically generate the required mesh annotations needed by the canonical mapping model for training and inference. To generate the synthetic dataset, we use a modified version of BlenderProc and Blender as the backend. This step generates an arbitrary number of images and corresponding image-level annotations. The resulting dataset is used to train a canonical mapping model that is based on the CSE approach. Finally, the inferred pixel-to-vertex matches are converted to task-specific representations. The qualitative results of the trained model and conversion algorithm on the real images are presented in Fig. \ref{fig:test_results}.

\subsection{Dataset generation}
A dataset for the approach consists of two main parts: 3D mesh (Fig. \ref{fig:cm_overview}a) with mesh-level annotations and one or several image datasets with image-level annotations. The mesh generated by scanning the object with a smartphone app is intended to be used without any extensive manual processing. The only manual step applied to our test mesh was coordinate frame rotation to match our preferred convention. However, this step is optional and the frame can be automatically assigned, e.g., based on principal axes. Then, we automatically generate the following annotations for the mesh, in accordance with the description of annotations in CSE paper: Laplace-Beltrami operator (lbo) features, pairwise geodesic distances, and vertex symmetry along the principal axis (for data augmentation). 

To generate the image-level annotations, we use Blender as the backend for scene configuration and rendering and a modified version of BlenderProc library to generate custom annotations. First, we define a Blender scene by using a generic warehouse environment texture and placing a table (Fig. \ref{fig:cm_overview}b). Then, to generate each training example, we set a random light source location and intensity, a random camera pose (restricted to a range above the table), and place the object in a random pose on the table. Swapping of table textures and adding clutter around the object is done to improve the generalization of the trained model to the real images. Each rendered image has the following corresponding annotations for each target object: bounding box, segmentation mask, identifier for the corresponding object mesh, $N$ keypoints each containing $xy$ pixel coordinate and the index of a corresponding vertex in the mesh. To obtain the latter annotation, we use ray casting to determine which of the vertices of the mesh are projected on a given pixel based on the current mesh and camera poses. It should be noted that the entire scene definition is done using the Python API, and a generic scene (i.e. object on a table) defined once can be reused for arbitrary objects just by swapping the mesh.

We also provide an interactive interface (Fig. \ref{fig:cm_overview}d) to allow a user to select vertices of interest that can be used to convert canonical mapping to a task-specific descriptor (usually, 6D Pose). The user is presented with a mesh and is able to click on any vertex to designate it as a seed. Using an adjustable geodesic distance threshold, neighboring vertices are grouped together around the seed. This method allows for defining an object part or a surface point (e.g. a grasping point), which is treated as a special case of a part. In both cases, the centroid of the grouped vertices is used to define the origin of the part and its orientation is set to match that of the parent object.

\subsection{Canonical Mapping model}
Continuous Surface Embeddings method works by predicting an embedding for each pixel of the object and then matching those embeddings to precomputed vertex-level surface features based on their Euclidean distance. We use 10K automatically generated synthetic training examples to train a category-specific CSE model. The model was trained from scratch (except for the Renset50-FPN backbone that was initialized with ImageNet-based weights) following the original paper's protocol for training a Human category model. Since the CSE architecture is based on Mask \mbox{R-CNN}, the model is capable to localize and segment objects in the scene prior to computing the canonical mapping (Fig. \ref{fig:cm_overview}h depicts colorized mapping visualized on the detected and segmented object). After training on several object categories, we empirically established that the default hyperparameters work well for a wide range of objects. The lack of intricate hyperparameter tuning further simplifies our pipeline since the training step on generated data can be performed automatically. Our robotic experiments were performed with one of the trained objects: a soft toy turtle. The reference mesh for the object  to which pixels are mapped consists of around 25K vertices. The availability of accurate dense mapping makes the mesh itself a powerful representation for robotic manipulation as grasping points, desired orientations, trajectories, or even gripper preferences can be defined with respect to the mesh and then seamlessly transferred to the observed object instance using canonical mapping.

\begin{algorithm}[b!]
\caption{Compute 6D Poses from Canonical Mapping}
\label{algo:task_conversion}
\SetAlgoLined
\SetKwInOut{Input}{Input}
\SetKwInOut{Output}{Output}

\begin{flalign*}
\textbf{In \hspace{0.5mm}: } & \text{Unprojected foreground pixels: } & \{x_0, \ldots, x_n\}\\
& \text{Vertices of canonical mesh: } & \{v_0, \ldots, v_m\}\\
& \text{Predicted pixel embeddings: } & E\in \mathbf{R}^{n\times d}\\
& \text{Pre-computed vertex embeddings: } & \hat{E}\in \mathbf{R}^{m\times d}\\
\textbf{Out: } & \text{Array of 6D Poses: }&\\
& \{Rt, Rt_{part_1}, Rt_{part_2}, \ldots\}, & Rt\in  \mathbf{R}^{4\times 4}
\end{flalign*}

\textbf{1}. Build a euclidean distance matrix for pixel embedding vectors and their top $K$ closest vertex embedding vectors:
\begin{fleqn}
\[
k_{i,j}\in \{0,\ldots, m\}, \quad i\in n, j\in \{0,\ldots,K\}
\]
\end{fleqn}
\begin{fleqn}
\[
\underset{n\times K}{D}=
\left[\begin{matrix}
\lVert E_0-\hat{E}_{k_{0,0}} \rVert & \dots & \lVert E_0-\hat{E}_{k_{0,K}} \rVert \\
\dots & & \dots & \\
\lVert E_n-\hat{E}_{k_{n,0}} \rVert & \dots & \lVert E_n-\hat{E}_{k_{n,K}} \rVert
\end{matrix}\right] 
\]
\end{fleqn}

\textbf{2}. Create a mask to filter out bad matches and outliers:
\begin{fleqn}
\[M^0, M^1\in \mathbf{R}^{n\times K},  \forall i \in n, \forall j\in \{0,\ldots,K\}\]
\[\textit{max\_dist: } \theta_0\in \mathbf{R}, \textit{outlier\_max\_dist: } \theta_1\in \mathbf{R}\]
\[
M^0_{i,j}= \left\{\begin{array}{lr}
        1, & D_{i,j} < \theta_0 \\
        0, & \text{else}
        \end{array}\right\} 
\]
\[
M^1_{i,j}= \left\{\begin{array}{lr}
        1, & (D_{i,j}-med(D_{i,0},\ldots ,D_{i,K}) ) < \theta_1\\
        0, & \text{else}
        \end{array}\right\} 
\]
\[
M_{i,j} = M^0_{i,j} \cdot M^1_{i,j}
\]
\end{fleqn}

\textbf{3}. Get unprojected pixels:
\begin{fleqn}
\[
X' = \{x_0, \ldots,x_n\}
\]
and the mean of corresponding filtered vertices:
\[
V' = \{\frac{v^T_{k_0}M_0}{\sum{M_0}},\ldots,\frac{v^T_{k_n}M_n}{\sum{M_n}}\}
\]
\end{fleqn}

\textbf{4}. Use Least-Squares to estimate best-fit transform (Pose):
\begin{fleqn}
\[
\text{Solve for } Rt\in \mathbf{R}^{4\times 4}: V' = Rt X'
\]
\end{fleqn}

\textbf{5}. Repeat steps 3 and 4 keeping only manually selected vertex subsets to compute part poses.

\end{algorithm}

\subsection{Task-specific conversion}
Many manipulation tasks benefit from having a more generic and compact object representation, such as 6D Pose. As previously discussed, there are many established methods that can predict discrete output such as pose, however, they need to be trained specifically for that task from the start. We demonstrate how canonical mapping can be converted to the object's rigid 6D Pose, but also how vertex groups or grasping points defined on the mesh using the UI can be used to extract part poses and grasping vectors without any retraining (Fig. \ref{fig:cm_overview}i). The approach uses a distance matrix between pixel and vertex embeddings maintaining several possible vertex candidates for each pixel and filtering the outliers based on the maximum overall distance and the distance from the median of all candidates for a given pixel. The two sets of 3D correspondences are then obtained by (1) taking the mean of the remaining per-pixel vertex candidates and (2) unprojecting the pixels to compute the observed 3D points. The unprojection requires a distance estimate for a pixel, which we obtained using the depth image from the RGB-D sensor. However, the approximate distance value can be also estimated by enforcing the same pairwise distance between unprojected points as between their corresponding vertices. The overall conversion process is described in Algorithm \ref{algo:task_conversion}. An added benefit of the method (shared by other model-based approaches) is that because of the availability of the object's mesh, it can estimate the poses of parts that are occluded or not visible in a given image. Naturally, such a pose assumes rigidity and cannot estimate articulations of occluded parts.

\subsection{Robotic manipulation}

Robotic manipulation strategies vary from simple to exceptionally complex, and the methods employed are dependent on the grippers used and the object to be manipulated (e.g., \cite{babin2021mechanisms, li2019review}). However, the general philosophy is to create a robust algorithm that will reliably manipulate a product in a desired manner. Often, robotic manipulation will rely on fixturing for accuracy, such that the object is in a known location and orientation. In semi-structured environments, these assumptions do not hold. The manipulation strategy must be adaptable for these varying circumstances. For the purposes of this research, a very simple grasping strategy was used with the assumption that the information given by the canonical mapping algorithm will be sufficient to manipulate the object in any position and orientation within the workspace of the robotic manipulator.

This strategy assumed that the proposed algorithm given in Algorithm \ref{algo:task_conversion} could correctly and accurately identify relevant part poses of a given object. For the purposes of this research, it is also assumed that the object to be manipulated is deformable, that the object is isolated on a flat plane, and that the object is within reach of the first manipulator. The formation of the grasp pose is shown in the 4x4 homogeneous transform matrix in Eq. \ref{eq:R1}, where the subscripts $w$, $o$, and $g$ denote ``world," ``object," and ``grasp" frames respectively, $P$ is a projection operator, and $\vec{p}$ is a position vector. The world information is assumed known, and the object information is given by \ref{algo:task_conversion}.

\begin{equation}
    \vec{x}_g = \vec{x}_o-P_{\vec{z}_w}(\vec{x}_o)
 \end{equation}

\begin{equation}
    B_g = \begin{bmatrix}
        \vec{x}_g & \vec{z}_w \times \vec{x}_g & \vec{z}_w & \vec{p}_w\\
        0 & 0 & 0 & 1
\end{bmatrix}
\label{eq:R1}
\end{equation}

Knowing the pose of elements within the object is enough to develop a suitable grasping strategy. In this instance, two-fingered and four-fingered grippers will be used that have a specific opening and closing orientation. To best facilitate a successful grasp, the end effectors will use the pose formulated in Eq. \ref{eq:R1}, with the open and close direction aligned with the minor axis of the object (the $\vec{y_g}$-axis in this example). Representation of a successful grasp is shown in Fig. \ref{fig:Experiment1}c and Fig. \ref{fig:Experiment1}d.

\section{EXPERIMENTS}

\begin{figure}[t]
    \centering
    \vspace{6pt}
    \includegraphics[width=\linewidth]{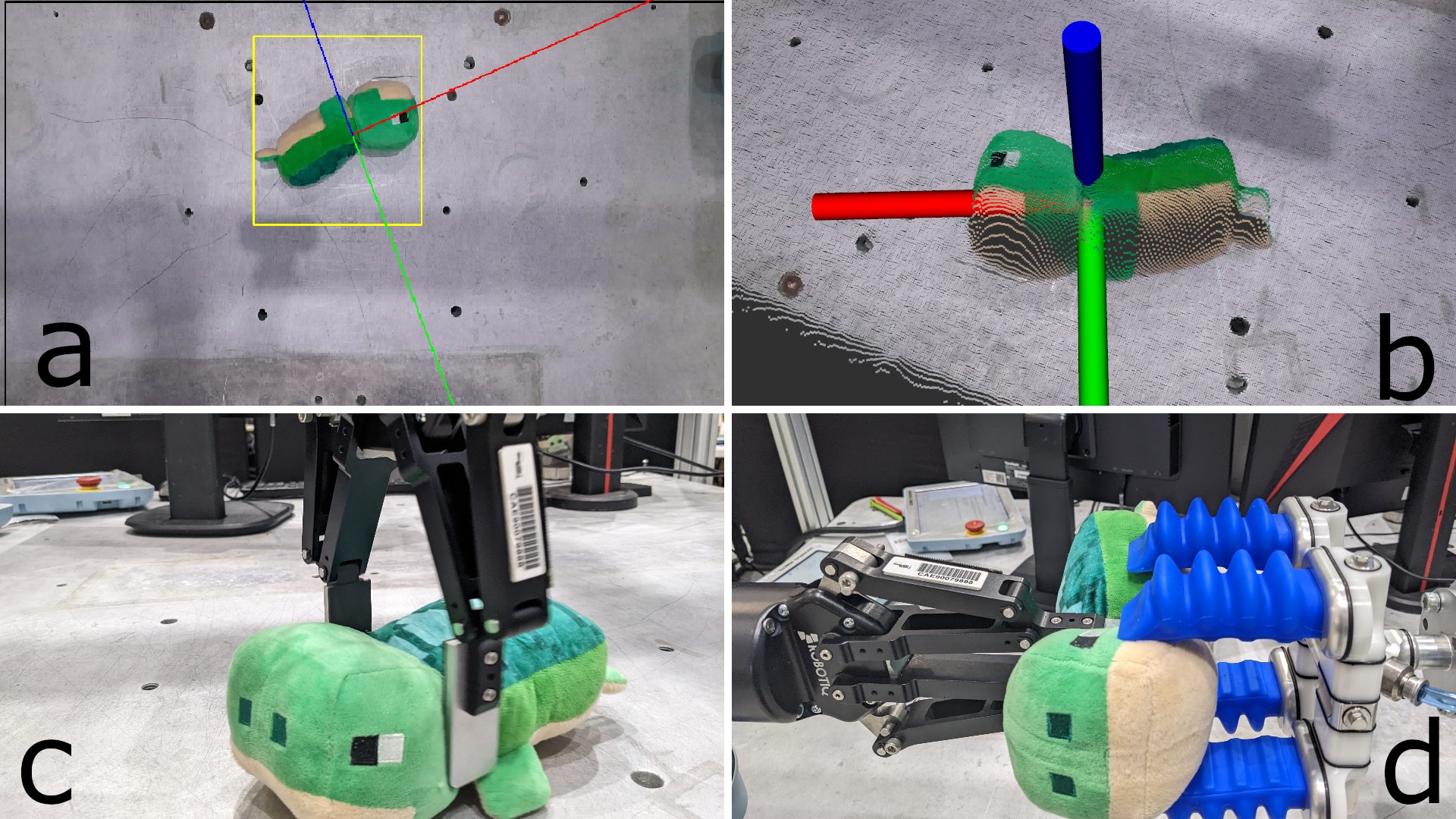}
    \caption{Representation of success metrics: (a) Segmentation of object in pixel space; (b) Estimating the pose of the object in Cartesian Space; (c) Grasping the object from pose information; and (d) a blind hand off to a second manipulator.}
    \label{fig:Experiment1}
\end{figure}

Proving the accuracy of a routine in an unstructured environment is challenging. Often, ground truth is unavailable or even unobtainable. Therefore, a series of experiments using the output from Algorithm \ref{algo:task_conversion} were applied to a robotic manipulation task to demonstrate the effectiveness of this perception methodology. This procedure consisted of two phases: grasping a deformable object using perception with one manipulator, and then a blind hand-off to a second manipulator. This technique allowed us to define four levels of accuracy spanning the domains of pixel space to Cartesian space, as shown in the following metrics:

\begin{itemize}
    \item M1: Segmentation of the object in pixel space, Fig. \ref{fig:Experiment1}a.
    \item M2: Estimating the pose of the object in Cartesian Space, Fig. \ref{fig:Experiment1}b.
    \item M3: Grasping the object from pose information with the first manipulator, Fig. \ref{fig:Experiment1}c.
    \item M4: Transferring the object to a second manipulator in a blind hand off, Fig. \ref{fig:Experiment1}d.
\end{itemize}

If the system is capable of all four of the above tasks, as seen in Fig. \ref{fig:Experiment1}, then the experiment is considered a complete success, and a robotic system has interacted with an unstructured environment with sufficient knowledge to perform a subsequent task without the need for additional information. 

We note that the grasping experiments were designed to evaluate the robustness and applicability of a canonical mapping descriptor generated for a novel object. While the experiments validate the canonical model's generalization from synthetic training data to real testing data, they don't evaluate category-level generalization to several object instances. However, encouragingly, the underlying canonical mapping approach, CSE, demonstrates high accuracy for several instance-rich animal categories using real-life annotated data. Additionally, the ease of scanning 3D meshes and automatic dataset generation used in our pipeline, make it reasonable to use several category instances to train the canonical model.

The hardware for this experiment consisted of two Universal Robots' UR5 model manipulators, a Robotiq two-fingered end effector, and a Soft Robotics end effector. The manipulators were placed on a rigid platform approximately 130 cm apart, and a Kinect Azure RGB-D camera was fixed approximately 80 cm above the platform facing down. For this experiment, a plush turtle toy was chosen as the test subject as it is deformable and semi-articulating. The following parts were selected and their poses generated using the UI tool: belly, back, left hand, right hand, left foot, right foot.

To test the effectiveness of the routine, four experiments were devised, as follows:

\begin{itemize}
    \item Experiment 1: Mid grab of an undeformed turtle plush toy.
    \item Experiment 2: Highest part grab of an undeformed turtle plush toy.
    \item Experiment 3: Mid grab of a deformed turtle plush toy.
    \item Experiment 4: Highest part grab of a deformed turtle plush toy.
\end{itemize}

\begin{figure}[t]
    \centering
    \vspace{6pt}
    \includegraphics[width=\linewidth]{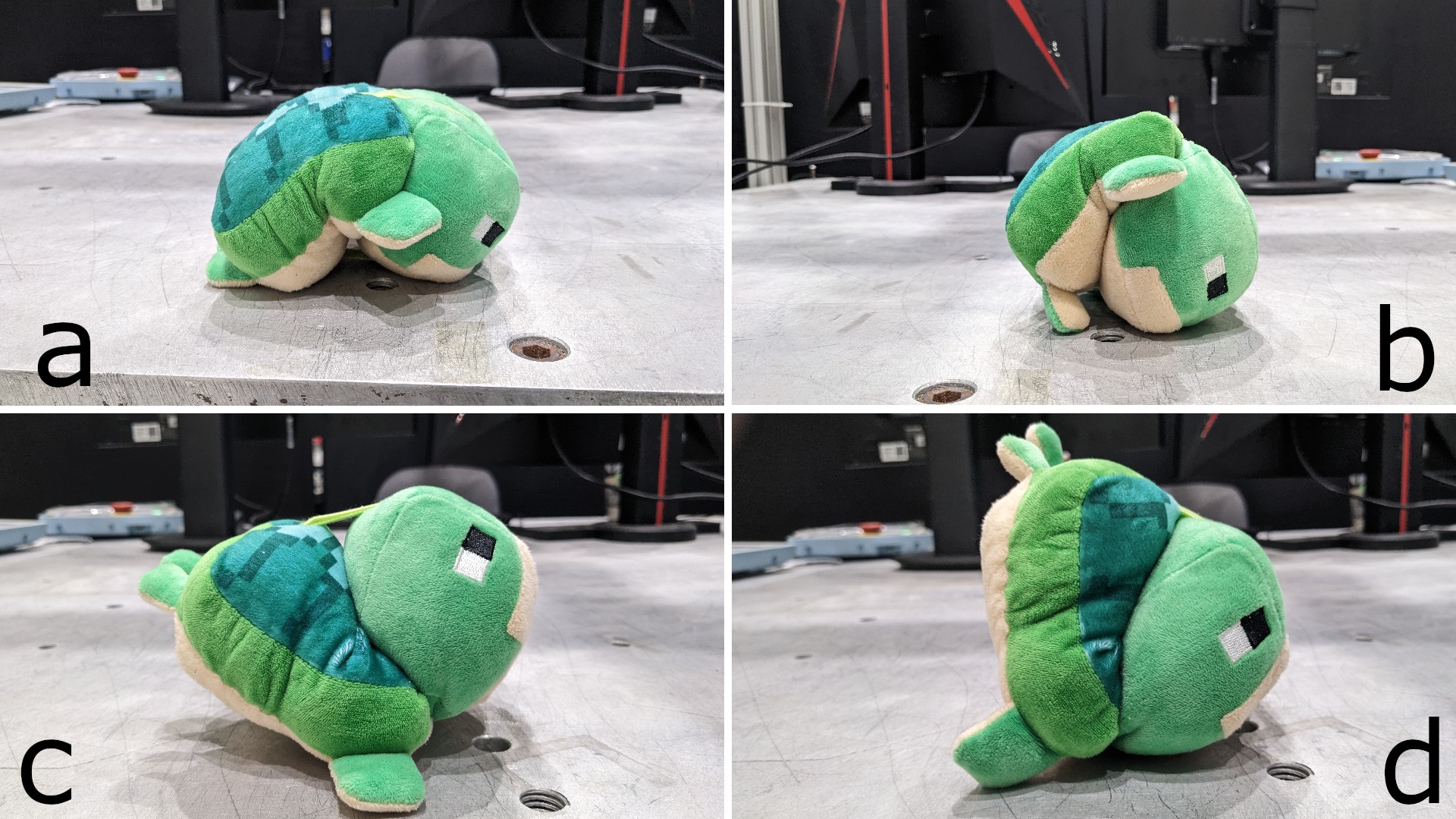}
    \caption{Example of deformed turtle toy designated as: (a) 90\textdegree \: belly, (b) 180\textdegree \: belly, (c) 90\textdegree \: back, and (d) 180\textdegree \: back.}
    \label{fig:Experiment2}
\end{figure}

These experiments break into two independent factors: the method of manipulation and the expected state of the object. The ``mid grab," as the name suggests, is a strategy where the gripper attempts to pinch the toy at approximately the mid-point of the object. This point was chosen as an average of the pose information given by the belly, back, left hand, and right hand. This area roughly corresponds to the middle of the toy and offers a good spot to routinely grab. The ``highest part grab'' dynamically identifies a part of the toy that is highest from the ground plane and attempts to grab that part. The ``undeformed" designation indicates that the toy was presented without external stresses. The ``deformed" toy, meanwhile, had a strap around the object folding it into one of four shapes, as seen in Fig. \ref{fig:Experiment2}. The results of these experiments are shown in Table \ref{tab:exp1}.

\begin{table}[!ht]
    \centering
    \begin{NiceTabular}{|l||l|l|l|l|l|}
    \hline
          & \textbf{Trial}  & \textbf{M1} & \textbf{M2} & \textbf{M3} & \textbf{M4}  \\ \hline \hline
        \multirow{5}{*}{\textbf{Experiment 1}} & Back & 100\% & 100\% & 100\% & 100\%  \\ \cline{2-6}
        & Belly & 100\% & 100\% & 100\% & 100\%  \\ \cline{2-6}
        & Left Side & 100\% & 100\% & 100\% & 100\%  \\ \cline{2-6}
        & Right Side & 100\% & 100\% & 100\% & 96\%  \\ \cline{2-6}
        & \textbf{Average} & \textbf{100\%} & \textbf{100\%} & \textbf{100\%} & \textbf{99\%}  \\ \hline \hline

         \multirow{5}{*}{\textbf{Experiment 2}} & Back & 100\% & 100\% & 100\% & 100\% \\ \cline{2-6}
        & Belly & 100\% & 100\% & 100\% & 100\% \\ \cline{2-6}
        & Left & 100\% & 100\% & 96\% & 84\% \\ \cline{2-6}
        & Right Back & 100\% & 100\% & 100\% & 100\% \\ \cline{2-6}
        & \textbf{Average} & \textbf{100\%} & \textbf{100\%} & \textbf{99\%} & \textbf{96\%} \\ \hline \hline

        \multirow{5}{*}{\textbf{Experiment 3}} & 90\textdegree \: Belly & 100\% & 100\% & 96\% & 64\%  \\ \cline{2-6}
        & 180\textdegree \: Belly & 100\% & 100\% & 92\% & 64\%  \\ \cline{2-6}
        & 90\textdegree \: Back & 100\% & 100\% & 92\% & 84\%  \\ \cline{2-6}
        & 180\textdegree \: Back & 100\% & 100\% & 76\% & 68\%  \\ \cline{2-6}
        & \textbf{Average} & \textbf{100\%} & \textbf{100\%} & \textbf{89\%} & \textbf{70\%}  \\ \hline \hline

        \multirow{5}{*}{\textbf{Experiment 4}} & 90\textdegree \: Belly & 100\% & 100\% & 92\% & 72\% \\ \cline{2-6}
        & 180\textdegree \: Belly & 100\% & 100\% & 92\% & 84\% \\ \cline{2-6}
        & 90\textdegree \: Back & 100\% & 100\% & 100\% & 72\% \\ \cline{2-6}
        & 180\textdegree \: Back & 100\% & 100\% & 92\% & 80\% \\ \cline{2-6}
        & \textbf{Average} & \textbf{100\%} & \textbf{100\%} & \textbf{94\%} & \textbf{77\%} \\ \hline
        
    \end{NiceTabular}
    
    \caption{\label{tab:exp1}Success rating for the various trials of deformed and undeformed toys using different grasping strategies. Each trial was run 25 times for a total of 400 attempts.}
\end{table}

These results show that, in every instance of the four hundred trials, the perception strategy successfully segmented the turtle from the surrounding pixels and correctly estimated the pose of the turtle to within the bounds of the physical model. The first experiment of a ``mid grab" of an undeformed toy was nearly one hundred percent successful. In the one instance of failure, the first manipulator knocked the toy over as it travelled to the pick point, which allowed for a successful grasp of the first gripper, but the trial failed the hand off. This result contrasts with the outcome of Experiment 2. In this experiment, the first manipulator failed to grasp the turtle once, and then it failed to hand off the turtle four times. The cause of these failures was the articulation of the toy's arm. The arm was identified as the ``highest part" and was chosen for grasping, but several times when it was grasped sufficiently for manipulation, the articulation caused a bend in the toy that then led to a failure in the hand off. Perhaps more interesting are the results from Experiments 3 and 4. In some of these trials, the turtle was deformed into an awkward shape that made it more difficult to successfully grasp and hand off according to these algorithms. For Experiment 3, these failures were due to the difficulty in reliably grasping the midsection when the system was deformed into an odd shape. However, Experiment 4 demonstrated that if the ``part" could be dynamically chosen, then it had an overall higher success rating in grasping and manipulating the object. Amongst all the trials, when the turtle was undeformed, the total success rating was greater than 97\%, and if the turtle was deformed, the success rating was approximately 73\%.

The intention of this research was not to design a grasping strategy that would prove to be 100\% effective. Rather, these trials demonstrate that Algorithm \ref{algo:task_conversion} is sufficiently advanced and provides enough reliable pose data that even simple grasping strategies have a high chance of success. Furthermore, allowing an algorithm to chose how it will manipulate an object based on the available data and inferred poses of the target object increases the chance for success. The information that the canonical mapping methodology provides is the real source of power that enables robust grasping algorithms.

\section{CONCLUSIONS}

This paper demonstrates that canonical mapping can serve as a near-universal descriptor for rigid, articulating, and deformable objects with powerful applications in robotic manipulation. Progress in photogrammetry and synthetic data generation allows a robust model for a novel physical object to be trained using an almost completely automated pipeline. After the training, the user can effortlessly adapt the model to traditional representations such as 6D Poses for any of the object's parts. Ease of training and suitability for a variety of objects, make the approach a technology capable of enabling new robotic applications in traditionally challenging environments.




\section*{ACKNOWLEDGMENT}

The research presented in this paper was supported by the Agricultural Technology Research Program of the Georgia Tech Research Institute.


\printbibliography 

\end{document}